# Variance-Based Rewards for Approximate Bayesian Reinforcement Learning

**Jonathan Sorg**
Computer Science & Engineering
University of Michigan

**Satinder Singh**
Computer Science & Engineering
University of Michigan

**Richard L. Lewis**
Department of Psychology
University of Michigan

## Abstract

The explore–exploit dilemma is one of the central challenges in Reinforcement Learning (RL). Bayesian RL solves the dilemma by providing the agent with information in the form of a prior distribution over environments; however, full Bayesian planning is intractable. Planning with the mean MDP is a common myopic approximation of Bayesian planning. We derive a novel reward bonus that is a function of the posterior distribution over environments, which, when added to the reward in planning with the mean MDP, results in an agent which explores efficiently and effectively. Although our method is similar to existing methods when given an uninformative or unstructured prior, unlike existing methods, our method can exploit structured priors. We prove that our method results in a polynomial sample complexity and empirically demonstrate its advantages in a structured exploration task.

## 1 Introduction

One of the central challenges of reinforcement learning (RL) is the explore–exploit dilemma. An agent must maximize its rewards (exploit) while simultaneously sacrificing immediate gains to learn about new ways to exploit in the future (explore). In one approach to addressing the dilemma, Bayesian Reinforcement Learning, the agent is endowed with an explicit representation of the distribution over the environments it could be in. As it acts and receives observations, it updates its belief about the environment distribution accordingly. A Bayes-optimal agent solves the explore–exploit dilemma by explicitly including information about its belief in its state representation and incorporating information changes into its plans (Duff, 2003). However, Bayes-optimal planning is intractable in general. A number of recent methods have attempted to approximate Bayesian planning (Poupart et al., 2006; Asmuth et al., 2009; Kolter and Ng, 2009), but this remains a challenging problem.

Another approach to addressing the explore–exploit dilemma is the explicit modification of the *objective* reward function—adding a reward bonus for exploration. We will refer to a modified reward function as an *internal* reward function. An agent that always exploits the internal reward function accomplishes both exploration and exploitation (with respect to the original objective reward function), thus solving the dilemma. This approach is (approximately) exemplified by many methods in the PAC framework (Kearns and Singh, 2002; Strehl et al., 2006; Strehl and Littman, 2008; Kolter and Ng, 2009) which bound the complexity of learning an MDP by explicitly motivating the agent to sample state-action pairs enough times to ensure it has explored sufficiently. Internal-reward methods have one advantage over the Bayesian approach—modifying the reward function can greatly influence behavior, often without greatly affecting computational cost.

In contrast, an important advantage of the Bayesian approach is that exploration is guided by prior knowledge of the environment. Some environments may require more exploration than others; some areas of the state space may be more uncertain than others; and most interestingly, information gained in one area of the state space may affect knowledge about other areas. The Bayesian approach expresses all of these through the specification of the agent's prior belief.

In this work, we contribute an internal-reward method for efficient exploration that takes advantage of prior knowledge in the form of a Bayesian prior. Thus our method retains the computational advantage of internal reward methods while obtaining guidance from a Bayesian prior. Specifically, our method provides a reward bonus in proportion to the square root of the variance of the agent's posterior distribution over envi-ronments. Our proposed method is similar to existing methods when using an uninformative prior, such as the independent Dirichlet distribution over transition dynamics. For structured priors however, we show that our method is capable of achieving a theoretically lower sample complexity bound than existing methods. We demonstrate that our method compares favorably to existing approximate Bayesian methods with both structured and unstructured priors in two environments, including the Hunt the Wumpus Environment (Russell and Norvig, 2002), where prior knowledge is critical and over-exploration can lead to permanent death.

## 2 Method

An MDP is the tuple $\langle S, A, R_\theta, P_\theta, \gamma \rangle$, where $S$ is the state space, $A$ is the action set, $R_\theta(s,a) \in [0,1]$ is the expected reward function over state–action pairs, $P_\theta(s'|s,a)$ is the probability of transitioning to state $s'$ given that action $a$ was taken in state $s$, and $\gamma$ is the discount factor. We often directly refer to an MDP's model parameters $\theta$ in the subscripts of the MDP's functions. Sometimes, when it is clear from context, we will use $S$ and $A$ to also refer to the cardinality of the corresponding sets.

### 2.1 Bayes-Optimal Planning

Unlike the standard RL setting, in Bayesian RL the agent is provided a-priori with the distribution of environments it could face, called the *prior* or the initial *belief* $b_0$. For belief $b$, we denote the probability of a particular MDP $\theta$ as $b(\theta)$. When the agent begins acting, it does not know which specific environment it is in, but instead updates its belief based on experience. During planning, a Bayes-optimal agent considers the effects of its own future changes in belief in addition to changes to physical state. We refer to this joint system as the information-state MDP. The full information state is the pair $\langle s, b \rangle$, where $s$ is the MDP state and $b$ is the belief state. After observing a state transition from $s$ to $s'$ and reward $r$, the agent updates its belief $b$ to the Bayesian *posterior* belief $b'$ using Bayes' rule. Therefore, a state transition in the information-state MDP is from $\langle s, b \rangle$ to $\langle s', b' \rangle$.

Let $B$ be the set of possible belief states. We define the *mean reward function* given belief $b$ as $R_b(s,a) \stackrel{\text{def}}{=} \int_\theta R_\theta(s,a)b(\theta)$. We similarly define the *mean transition function* as $P_b(s'|s,a) \stackrel{\text{def}}{=} \int_\theta P_\theta(s'|s,a)b(\theta)$. The optimal behavior in the information state MDP is computed by solving the Bellman optimality equations: $\forall \langle s,b \rangle \in S \times B, a \in A$,

$$Q^*(\langle s,b\rangle, a) = R_b(s,a) + \gamma \sum_{s'} P_b(s'|s,a) V^*(\langle s',b'\rangle),$$

where $V^*(\langle s',b'\rangle) \stackrel{\text{def}}{=} \max_a Q^*(\langle s',b'\rangle, a)$. The function $Q^*(\langle s,b\rangle, a)$ is the Bayes-optimal action-value function. An agent that acts greedily with respect to $Q^*(\langle s,b\rangle, \cdot)$ acts Bayes-optimally.

Full Bayesian planning (i.e., solving the Bellman optimality equations above for the Bayes-optimal action-value function) is expensive, because for many priors, the set of possible belief states $B$ is large or infinite. Furthermore, because the agent is constantly learning and updating its belief, an agent may rarely or never (depending on the prior) revisit belief states. For these reasons, agents must approximate Bayes optimality in general.

### 2.2 Mean MDP + Reward Bonus

Planning with the mean MDP with respect to belief $b$ is a simple, myopic approximation of Bayesian planning which removes the state-space explosion of belief states while preserving physical state dynamics. The resulting Bellman equations are identical to the above equations for Bayesian planning, except the belief state is not updated on the right-hand side: $\forall s \in S, a \in A$,

$$Q_b^*(s,a) = R_b(s,a) + \gamma \sum_{s'} P_b(s'|s,a) V_b^*(s'), \quad (1)$$

where $V_b^*(s') \stackrel{\text{def}}{=} \max_a Q_b^*(s',a)$. We denote the value's dependence on the current belief $b$ in the subscript to emphasize the belief's invariance during planning. At each time step, the Mean MDP (MMDP) agent acts greedily with respect to $Q_b^*$. After observing the result of its action, it updates its belief to $b'$ using Bayes' rule and then computes $Q_{b'}^*$. Thus, an agent which plans in the mean MDP *does* update its belief as it receives experience, but does not do so during planning.

We will use a reward bonus to help compensate for the information not accounted for in the mean MDP planning approximation (cf. Equation 1) to full Bayesian planning. Such a Mean MDP plus Reward Bonus (MMDP+RB) agent is identical to the mean MDP agent described above, except it uses an internal reward function defined as the mean reward function plus an added reward bonus term: $\tilde{R}_b(s,a) = R_b(s,a) + \hat{R}_b(s,a)$. Formally, an MMDP+RB agent with belief $b$ always acts greedily with respect to the action-value function defined by: $\forall s \in S, a \in A$,

$$\tilde{Q}_b^*(s,a) = \tilde{R}_b(s,a) + \gamma \sum_{s'} P_b(s'|s,a) \tilde{V}_b^*(s'), \quad (2)$$

where $\tilde{V}_b^*(s') \stackrel{\text{def}}{=} \max_a \tilde{Q}_b^*(s',a)$.

The approximate Bayesian algorithm of Kolter and Ng (2009) is an existing algorithm that is of the MMDP+RB form, albeit in a limited special case. Let

$n_{s,a}$ be the number of times that state–action pair $(s, a)$ has been sampled. They showed that a MMDP+RB agent with a reward bonus of $\hat{R}_b(s, a) = \beta/n_{s,a}$ approximates Bayesian planning in polynomial time with high probability (for appropriate choice of constant $\beta$), in the special case of an independent Dirichlet prior over transition dynamics per state–action pair and a known reward function. We will use this algorithm as one baseline in our experiments.

The MBIE-EB algorithm (Strehl et al., 2006) has the same form as MMDP+RB, though it is not derived in a Bayesian setting. In this algorithm, the agent plans in the maximum-likelihood estimate of the MDP; this is very similar to the mean MDP of a Dirichlet prior. MBIE-EB features a reward bonus of $\hat{R}_b(s,a) = \beta/\sqrt{n_{s,a}}$. As described in detail later, this algorithm also connects closely to our method; we will use it as another baseline in our experiments.

Both of these baselines share the property that their reward bonuses decrease independently per state–action pair as each is sampled. Both intuitively measure the uncertainty the agent has for that state–action pair. However, neither accounts for information contained in the prior distribution (unless that prior is a factored Dirichlet). In the next subsection, we define the variance-based reward bonus, which is capable of measuring the uncertainty of arbitrary Bayesian priors over environments.

### 2.3 Variance-Based Reward Bonus

The variance of the model parameters are defined as:

$$\sigma^2_{R_b(s,a)} \stackrel{\text{def}}{=} \int_\theta R_\theta(s,a)^2 b(\theta) - R_b(s,a)^2, \text{ and}$$

$$\sigma^2_{P_b(s'|s,a)} \stackrel{\text{def}}{=} \int_\theta P_\theta(s'|s,a)^2 b(\theta) - P_b(s'|s,a)^2.$$

Importantly, these are not the variance of the world dynamics but instead are the variance of the model parameters with respect to the agent's belief. As the agent gathers experience in the world—as the agent becomes more certain of the truth—these values will tend to decrease to 0, regardless of how stochastic the world is. Notice that we have made no state–action independence requirements on the prior. Therefore, depending on the belief, experience gained in one state may affect the variance term in any other state.

We define the variance reward bonus using these variance terms:

$$\hat{R}_b(s,a) \stackrel{\text{def}}{=} \beta_R \sigma_{R_b(s,a)} + \beta_P \sqrt{\sum_{s'} \sigma^2_{P_b(s'|s,a)}},$$

for some constants $\beta_R$ and $\beta_P$. Although the precise form of this reward may seem unintuitive, in the next section, we will show that this reward bonus can be used to bound the error of the mean MDP, with respect to the random (drawn from the prior) true MDP with high probability. Using this fact, we will show that there exist constants $\beta_R$ and $\beta_P$ such that an agent that acts greedily with respect to $\tilde{Q}_b^*(s,a)$ acts optimally with respect to the random true MDP, for all but a polynomially bounded number of time steps with high probability.

## 3 Sample Complexity

The key insight behind the use of variance as a measure of the agent's uncertainty is that we can bound the deviation of the mean MDP model from the true MDP using Chebyshev's inequality, which states that the deviation of a random variable from its mean is no more than a multiple of its variance, with high probability: $\Pr(|X - \mu| \geq \eta) \leq \frac{\sigma^2}{\eta^2}$, where $X$ is a random variable, $\mu$ and $\sigma$ are its mean and standard deviation, and $\eta$ bounds the deviation from the mean.

**Lemma 1.** *For any belief $b$ and any state–action pair $(s, a)$, there exists a value $\eta_P(s, a, b)$ which bounds the max-norm error of the mean transition model with probability at least $1 - \rho$:*

$$Pr\left[\|P_\theta(\cdot|s,a) - P_b(\cdot|s,a)\|_\infty < \eta_P(s,a,b)\right] > 1 - \rho,$$

*where $\|\cdot\|_\infty$ is the max norm, and this inequality is satisfied if*

$$\eta_P(s,a,b) = \frac{1}{\sqrt{\rho}} \sqrt{\sum_{s'} \sigma^2_{P_b(s'|s,a)}}.$$

*Proof.* We apply the union bound over next-states.

$$Pr\left[\|P_\theta(\cdot|s,a) - P_b(\cdot|s,a)\|_\infty \geq \eta_P(s,a,b)\right]$$
$$\leq \sum_{s'} Pr\left[|P_\theta(s'|s,a) - P_b(s'|s,a)| \geq \eta_P(s,a,b)\right]$$
$$\leq \sum_{s'} \sigma^2_{P_b(s'|s,a)}/\eta_P^2(s,a,b) \leq \rho.$$

Solving for $\eta_P(s, a, b)$ completes the proof. □

We can do a similar analysis for the reward function, though it does not require the union bound.

**Lemma 2.** *For any belief $b$ and any state–action pair $(s, a)$, if the belief distribution over rewards has finite variance, then there exists a value $\eta_R(s, a, b)$ which bounds the error of the mean reward function with probability at least $1 - \rho$:*

$$Pr\left[|R_\theta(s,a) - R_b(s,a)| < \eta_R(s,a,b)\right] > 1 - \rho.$$

*and this is satisfied if $\eta_R(s,a,b) = \sigma_{R_b(s,a)}/\sqrt{\rho}$.*

**Algorithm 1**: Bounded Variance Reward Algorithm
---
**Input**: $s_0$, $b_0$, $\beta_R$, $\beta_P$, $C$
$\forall s, a \quad c_{s,a} \leftarrow 0$, $known_{s,a} \leftarrow false$
$\forall s, a \quad Q(s, a) \leftarrow \tilde{Q}^*_{b_0}(s, a)$
**for** $t \leftarrow 0, 1, 2, 3, \ldots$ **do**
  $a_t \leftarrow \arg\max_a Q(s_t, a)$
  $r_t, s_{t+1} \leftarrow \texttt{takeAction}(a_t)$
  $b_{t+1} \leftarrow \texttt{updateBelief}(b_t, s_t, a_t, r_t, s_{t+1})$
  $c_{s_t, a_t} \leftarrow c_{s_t, a_t} + 1$
  **if** $\neg known_{s_t, a_t} \land (c_{s_t, a_t} \geq C(s_t, a_t, b_0, \epsilon, \delta))$ **then**
    $\forall s, a \; Q(s, a) \leftarrow \tilde{Q}^*_{b_{t+1}}(s, a)$
    $known_{s_t, a_t} \leftarrow true$;
  **end**
**end**

---

Although the variance reward MMDP+RB agent was described in detail in Section 2.2, we analyze a slightly different agent in our theoretical analysis, presented in Algorithm 1. It differs in only one significant manner. Algorithm 1 takes as input a sample complexity parameter $C(s, a)$, which is a prior-dependent term indicating the number of times state–action pair $(s, a)$ must be sampled before it becomes *known*. Algorithm 1 only updates its value function estimate each time a state–action pair becomes known. This allows us to bound the number of times the agent plans.

In this section, we bound the sample complexity of Algorithm 1, following the abstract PAC framework by Strehl et al. (2006). A central aspect of this framework is the principle of optimism in the face of uncertainty. By ensuring the agent remains optimistic, we can ensure that the agent does not ignore potentially lucrative opportunities. The model error bounds above allow us to provide a reward bonus that ensures optimism. Unless stated otherwise, the proofs in this section are deferred to the Appendix.

**Lemma 3** (Optimism). *Let the reward bonus be $\hat{R}_b(s, a) = \frac{1}{\sqrt{\rho}} \left( \sigma_{R_b(s,a)} + \frac{\gamma S}{1 - \gamma} \sqrt{\sum_{s'} \sigma^2_{P_b(s'|s,a)}} \right)$, then the value function computed by Algorithm 1 is optimistic with probability at least $1 - 2S^2 A^2 \rho$ for every planning step during its execution.*

The convergence rate of Algorithm 1 depends on the convergence rate of the posterior distribution. Here, we abstractly define this rate and allow it to depend on state and action. We will later provide examples of this function for specific classes of priors. Note that because we have defined the reward bonus in such a way that it is an upper bound on the error of the mean MDP, defining the sample complexity with respect to the reward bonus term bounds the number of samples before we have an accurate model.

**Definition 1.** Define the sample complexity function, $f(b_0, s, a, \epsilon, \delta, \rho)$, as the minimum number $c$ such that if $d > c$ transitions from $(s, a)$ have been observed, starting from belief $b_0$, the additive reward term $\hat{R}$ using the updated belief $b_d$ is less than $\epsilon$, i.e., $\hat{R}_{b_d}(s, a) = \frac{1}{\sqrt{\rho}} \left( \sigma_{R_{b_d}(s,a)} + \frac{\gamma S}{1-\gamma} \sqrt{\sum_{s'} \sigma^2_{P_{b_d}(s'|s,a)}} \right) < \epsilon$, with probability at least $1 - \delta$. We will refer to a state–action pair as *known* if it has been sampled $c$ or more times.

We make one important assumption about $f$: experience gained by sampling one state–action pair does not increase the sample complexity of another state–action pair. This assumption is trivially true when the prior is independent per state–action pair. Furthermore, we expect this to be a reasonable assumption in many correlated priors. In fact, we believe our method's performance on correlated priors to be one of its strengths. Rather than hurting convergence, experience gained from one area of the state space should in general reduce the number of samples required from another, and we present evidence consistent with this below.

Next, we present our central theoretical result that bounds the sample complexity with respect to the true MDP. This is distinct from the sample complexity bound of Kolter and Ng (2009) which is defined with respect to the information state MDP.

**Theorem 1.** *Let the sample complexity of state $s$ and action $a$ be $C(s, a) = f(b_0, s, a, \frac{1}{4}\epsilon(1 - \gamma)^2, \frac{\delta}{SA}, \frac{\delta}{2S^2 A^2})$. Let the internal reward $\hat{R}_b$ be defined as in Lemma 3 with $\rho = \frac{\delta}{2S^2 A^2}$. Let $\theta^*$ be the random true model parameters distributed according to the prior belief $b_0$. Algorithm 1 will follow a $4\epsilon$-optimal policy from its current state, with respect to the MDP $\theta^*$, on all but*

$$O \left( \frac{\sum_{s,a} C(s, a)}{\epsilon(1 - \gamma)^2} \ln \frac{1}{\delta} \ln \frac{1}{\epsilon(1 - \gamma)} \right) \quad (3)$$

*time steps with probability at least $1 - 4\delta$.*

Theorem 1 can be applied to many prior distributions. In the remainder of this section, we apply it to two simple special cases. First, we provide a concrete bound in the case of an independent Dirichlet prior and a known reward function. We use this special case to connect our result to related work.

**Lemma 4.** *(Independent Dirichlet Prior) Let $n_{s,a}$ be the number of times state–action pair $(s, a)$ has been sampled. For a known reward function and an independent Dirichlet prior over next-state transition dynamics for each state–action pair, the internal reward feature $\eta_P(s, a, b)$ decreases at a rate of $O(1/\sqrt{n_{s,a}})$.*

*Proof.* The square root of the sum of the variance terms

for the Dirichlet distribution is

$$\sqrt{\sum_{s'} \sigma^2_{P_b(s'|s,a)}} = \sqrt{\frac{\sum_{s'} P_b(s'|s,a)(1 - P_b(s'|s,a))}{n_{s,a} + 1}}$$
$$\leq 1/\sqrt{n_{s,a} + 1}. \qquad \square$$

Lemma 5, which we present without proof, uses the result in Lemma 4 to provide a sample complexity bound for Algorithm 1 with a Dirichlet prior.

**Lemma 5.** *The sample complexity function for an independent Dirichlet prior over transition dynamics and a known reward function is $f(b_0, s, a, \epsilon, \delta, \rho) = \gamma^2 S^2 / (\rho \epsilon^2 (1-\gamma)^2)$, where $S$ is the number of states.*

As stated before, the mean MDP for a Dirichlet prior is analogous to the MLE estimate MDP in MBIE-EB. Notice also that the $O(1/\sqrt{n_{s,a}})$ reward bonus derived here is similar to the reward bonus in MBIE-EB. In other words, we have effectively re-derived MBIE-EB using Bayesian methods; one could replace our variance term with a $1/\sqrt{n_{s,a}+1}$ reward bonus in our proofs and produce a similar result. Thus, our method is similar to existing methods when using the Dirichlet distribution, an unstructured and uninformative prior. That said, the variance term provides a slightly tighter upper bound than does $1/\sqrt{n_{s,a}+1}$, because it accounts for the distribution of observed data.

The advance of Algorithm 1 over prior methods lies in its ability to take advantage of structured and informative priors. As an initial and very simple example of this advantage, we present the sample complexity function for a stochastic prior over unknown deterministic MDPs. Utilizing this knowledge does require full probabilistic Bayesian reasoning—the effects of unknown actions will appear stochastic. However, after sampling a state–action pair once, the agent will know its effect. We state this formally as a lemma without proof.

**Lemma 6.** *(Prior over Deterministic MDPs) Let $b_0$ be a prior over deterministic worlds. The sample complexity function $f(b_0, s, a, \epsilon, \delta, \rho) \leq 1$.*

Furthermore, if the distribution is not independent between state–action pairs, sampling one state–action pair may cause the variance associated with other state–action pairs to also be set to 0. In fact, the variance feature for unobserved state–action pairs can sometimes increase with experience; however, the sample complexity result always holds.

## 4 Comparison Methods

Our method benefits from the Bayesian prior in two ways: (1) it uses the prior to generate the Mean MDP;

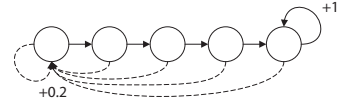

Figure 1: Chain Environment

(2) it uses the variance calculation to guide exploration. To properly demonstrate that the variance reward bonus deserves the credit for our method's success, all comparison methods will be given the same prior belief, and all will properly update their posterior belief given the prior.

The simplest baseline for our method to compete against is the mean MDP agent with no reward bonus. To fairly compare against MBIE-EB (Strehl and Littman, 2008) and the approximate Bayesian method of Kolter and Ng (2009), we test the corresponding MMDP+RB agents with reward bonuses of $O(1/\sqrt{n_{s,a}})$ and $O(1/n_{s,a})$, respectively.

The BOSS algorithm (Asmuth et al., 2009) is *not* an internal-reward approach, but it is a direct competitor to our method in another sense. It is the only other algorithm we are aware of which currently provides sample complexity guarantees as a function of a Bayesian prior. Each time it plans, it samples $K$ (a parameter) MDPs from the posterior distribution. It then plans in a combined MDP that has the same state space, but each state has $K \times A$ available actions. Essentially, the combined MDP allows the agent to choose, independently in each state, which sampled MDP's dynamics it would like to follow. This planning method is optimistic with enough samples $K$.

There are many other approximate methods for Bayesian planning. By using a standard benchmark task below, we are able to compare against the published results for one such method. The BEETLE algorithm (Poupart et al., 2006) directly approximates full Bayesian planning by compressing the information state space.

## 5 Empirical Results

In this section, we first compare our method on a standard benchmark problem, and then on a problem with an interesting structured prior.

### 5.1 Chain Environment

We will use the 5-state chain environment shown in Figure 1 to demonstrate two points. (1) For arbitrary priors in this task our proposed variance reward method compares favorably to other methods which approximate Bayesian planning. (2) When given an unstructured prior, such as a Dirichlet distribution, the performance of the variance reward bonus is similar to the performance of the $O(1/\sqrt{n_{s,a}})$ and $O(1/n_{s,a})$ reward bonuses, as discussed above.

Table 1: Chain Experiment Results

| Algorithm | Tied Prior | Semi Prior | Full Prior |
|---|---|---|---|
| BEETLE | 3,650 | 3,648 | 1,754 |
| BOSS | 3,657 | 3,651 | 3,003 |
| Mean | 3,642 | 3,257 | 3,078 |
| $O(1/n)$ | 3,645 | 3,642 | 3,430 |
| $O(1/\sqrt{n})$ | 3,645 | 3,642 | 3,462 |
| Variance Reward | 3,645 | 3,637 | 3,465 |

The chain environment has two actions: Action A (solid) advances the agent along the chain, and Action B (dashed) resets the agent to the first node. When taken from the last node, Action A leaves the agent where it is and gives a reward of 1—otherwise it results in 0 reward. Action B gives a reward of 0.2 in all states. However, with probability 0.2 the agent "slips" and the outcomes are switched. Optimal behavior always chooses Action A.

This environment was designed to require smart exploration, because the optimal policy produces distant reward while there are many sub-optimal policies which yield immediate reward. Past works (Poupart et al., 2006; Asmuth et al., 2009) consider the performance of agents with different priors in this environment. In the *Full* prior, the agent uses an independent Dirichlet prior distribution for each state–action pair. Under the *Tied* prior, the agent knows the underlying transition dynamics except for the value of a single slip probability that is shared between all state–action pairs. The *Semi*-tied prior allows for a different slip probability for each action which is shared across states. In Tied and Semi, the prior distribution over the slip probability is represented as a Beta distribution. In keeping with published results on this problem, Table 1 reports cumulative return in the first 1000 steps, averaged over 500 runs. Standard error is on the order of 20 to 50. The optimal policy for the true MDP scores 3677.

We present the performance of our method along side the comparison methods in Table 1. Each comparison algorithm is given the indicated prior and maintains the correct posterior distribution; however, each differs in its method used to approximate full Bayesian planning. Each of the reward-based methods is parameterized by a coefficient on the reward bonus. In Table 1, that coefficient was optimized separately for each result.

The reward bonus methods, including our method, perform as well as the other methods in general and outperform the alternatives in the case of the Full prior.

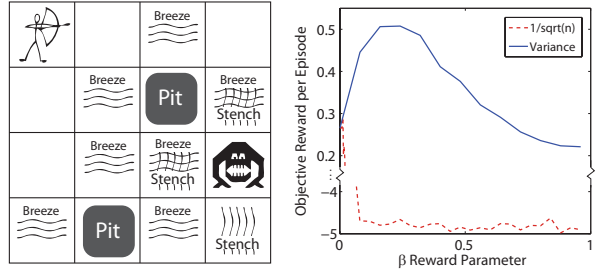

Figure 2: Hunt the Wumpus Environment

As predicted, the reward bonus methods perform similarly in the case of the Full (Dirichlet) prior. Although the Tied and Semi priors are structured, they essentially make the problem too easy—other than from the naive mean MDP approach, they fail to differentiate the methods.

### 5.2 Hunt the Wumpus

Next, we demonstrate the advantages of our method on a task with non-trivial correlated belief reasoning requirements, and in which poor early decisions can lead to the agent's death. The Hunt the Wumpus environment, adapted from Russell and Norvig (2002) and pictured in Figure 2, is a discrete world based on an old computer game which requires intelligent exploration. The world consists of a $4 \times 4$ cave. The agent always starts in the top-left corner. It can navigate by turning `left`, turning `right`, or moving `forward` one location. Lurking in the cave in a uniformly random location (other than the agent's starting location) is the wumpus, a beast that eats anyone who enters its location (ending the episode). The agent cannot see the wumpus, but if it is in a location cardinally adjacent to the wumpus, it can smell a vile stench. Each location also has a prior 0.2 probability of containing a deep pit that will trap a wandering adventurer (but not the wumpus), ending the episode. If the agent is next to any pit, it can feel a breeze, though it cannot sense in what direction the originating pit is. The agent carries a bow with one arrow and a has a `shoot` action. When executed, it fires an arrow the entire length of the cave in the direction it is facing. If it hits the wumpus, the wumpus dies and the agent receives 1 reward. Otherwise, the episode ends with 0 reward. At all other time steps, the agent receives a small penalty of -0.01 reward. In summary, the agent senses the tuple $\langle location, orientation, stench, breeze \rangle$. We represent the end conditions with two terminal states. In one, the agent receives 1 reward for killing the wumpus. In the other, the agent receives 0 for dying. We provide the agent with the true prior over environments.

Table 2: Hunt the Wumpus Results.

| | Parameter | Objective Reward/Episode |
|---|---|---|
| Variance | $\beta_P = 0.24$ | 0.508 |
| $1/n$ | $\beta = 0.012$ | 0.293 |
| $1/\sqrt{n}$ | $\beta = 0.012$ | 0.291 |
| BOSS | $K = 20$ | 0.183 |
| Mean MDP | N/A | 0.266 |

We evaluate the agents in a setting in which they have *one chance* to kill the wumpus—each episode is independently drawn from the prior. In other words, the pit and wumpus locations are resampled between episodes. Because the world's dynamics can be expressed as a function of the agent's observable location, and because they do not change during an episode, we can model each episode as a Bayesian distribution over MDPs. However, this environment is an unforgiving exploration benchmark; if the agent over-explores, it can fall into a pit or get eaten. There is no opportunity to learn from dying, because the agent only lives for one episode. In order to avoid likely death, while taking enough risks to find and kill the wumpus, an agent will need to properly utilize its prior.

In spite of the fact that the dynamics are deterministic, probabilistic Bayesian reasoning is necessary due to the stochastic nature of the prior. It is often the case that, given the agent's experience, some adjacent locations are more likely to contain pits than others. Information gathered in one location can increase or decrease the probabilities of the existence of pits or the wumpus in other locations. At times, the agent must take calculated risks—stepping into locations that have nonzero probability of containing a pit—in order to gain the information it needs to locate the wumpus.

We present the empirical results in Table 2. For each method, we present the mean objective reward obtained per episode, averaged over 500 episodes. Each episode is capped at a maximum length of 1,000 steps. As before, each reported method is given the same prior and properly updates its belief through experience; the algorithms differ in how they approximate Bayesian planning given those beliefs. Several of the algorithms have a free parameter which we optimized. For the internal reward methods, we searched over all reward bonus scalars $\beta$ in the range $[0, 0.04]$ in increments of $0.002$ and $[0.04, 1]$ in increments of $0.04$. For the variance method, we present $\beta_P$ only, because the reward function is known. For the BOSS method, we tested sample sizes ($K$) of 1,5,10,20,40, and 80.

The variance internal reward method achieves the top performance, because it follows an effective controlled exploration strategy. It is not optimal, however; because it enjoys exploring, it will occasionally spend time identifying the location of a few more pits after having already located the Wumpus. The BOSS agent performs poorly. As mentioned above, BOSS ensures optimism by building in the assumption that the agent knows, and in fact is in control of, which MDP the agent is in. Its control policy immediately turns and fires an arrow at time step 1—it chooses the imagined MDP in which the Wumpus is in its current line of sight. The mean MDP baseline agent performs better than BOSS, though its policy is merely a slightly better heuristic. It walks in a straight line until it encounters a breeze, at which point it fires in the direction of the most unexplored locations, unless it experiences a rare situation which disambiguates the wumpus's location, in which case it fires in the correct direction. The $O(1/n)$ and $O(1/\sqrt{n})$ reward bonus methods' policies are a small deviation from the mean MDP's policy—notice the small magnitude of the reward coefficient, just above the per-step penalty of the objective reward.

Figure 2(b) illustrates the effect of the choice of the reward scaling parameter $\beta$ on performance for both the variance and $O(1/\sqrt{n})$ reward methods (the graph is similar for $O(1/n)$). Note that the choice of $\beta = 0$ results in the mean MDP agent. As can be seen, the variance reward results in good performance for many choices of $\beta$, but $O(1/\sqrt{n})$ performs extremely poorly for large reward values—the agent spends the majority of its time following safe actions, such as turning in circles, that provide it with no information.

# 6 Conclusion

Although full Bayesian planning produces optimal behavior, it is intractable. In this work, we contributed a novel internal-reward algorithm with a sample complexity bound which derives its reward bonus from a Bayesian prior distribution. Our method is similar to existing approaches when given unstructured priors, such as the factored Dirichlet distribution; however, unlike previous reward bonuses, our approach is capable of exploiting structure in the prior. In addition to providing theoretical results supporting these claims, we demonstrated that our method exploits structured prior knowledge in the Hunt the Wumpus environment.

**Acknowledgements:** This work was supported by the Air Force Office of Scientific Research under grant FA9550-08-1-0418 as well as by NSF grant IIS 0905146. Any opinions, findings, conclusions, or recommendations expressed here are those of the authors and do not necessarily reflect the views of the sponsors.

## A  Technical Proofs

*Proof of Lemma 3.* Define the random variable $V^*_\theta(s)$ to be the value of state $s$ given the true model $\theta$. Let $\eta_P(s,a,b)$ and $\eta_R(s,a,b)$ be defined as in Lemmas 1 and 2. Given belief $b$, with probability at least $1 - 2SA\rho$,

$$V^*_\theta(s) = \max_a R_\theta(s,a) + \gamma \sum_{s'} P_\theta(s'|s,a) V^*_\theta(s')$$
$$\leq \max_a R_b(s,a) + \eta_R(s,a,b)$$
$$\quad + \gamma \sum_{s'} (P_b(s'|s,a) + \eta_P(s,a,b)) V^*_\theta(s')$$
$$\leq \max_a \tilde{R}_b(s,a) + \gamma \sum_{s'} P_b(s'|s,a) \tilde{V}^*_b(s') = \tilde{V}^*_b(s).$$

The first inequality is true with probability at least $1 - 2\rho$, and the final step can be shown by induction. This must be true for all state–action pairs, resulting in truth with probability greater than $1 - 2SA\rho$ by a union bound. For this to be true for the entire execution of Algorithm 1, it must be true for all value function updates, of which there are no more[1] than $SA$, resulting in the final bound of $1 - 2S^2A^2\rho$ through another application of the union bound. □

---

[1] Technically, including the initial planning phase, there are up to $SA + 1$ value updates.

**Definition 2.** Let $\theta = \langle S, A, P, R, \gamma \rangle$ be an MDP. Given a set of Q value estimates $Q(s,a)$ and a set $K$ of state–action pairs called the *known* state–action pairs, we define the *known state–action MDP* $\theta_K = \langle S \cup s_0, A, T_K, R_K, \gamma \rangle$ as follows. For each known state–action pair $(s,a) \in K$, $P_K(\cdot|s,a) = P(\cdot|s,a)$ and $R_K(s,a) = R(s,a)$. Define an additional absorbing state $s_0$ with $P_K(s_0|s_0, \cdot) = 1$ and $R_K(s_0, \cdot) = 0$. For each unknown state–action pair, the world deterministically transitions to $s_0$ ($\forall (s,a) \notin K, P(s_0|s,a) = 1$) with reward function $R_K(s,a) = Q(s,a)$.

Let $V_t(s) = \max_a Q_t(s,a)$ denote the agent's estimate of the value function at time $t$. Let $V^{\pi_t}_{\theta_{K_t}}$ denote the value of the agent's policy at time $t$ in the known state–action MDP defined with respect to the true MDP $\theta$ and the agent's value estimate $Q_t$.

**Lemma 7.** *If the state–action sample complexity in Algorithm 1 is $C(s,a) = f(b_0, s, a, \frac{1}{4}\epsilon(1-\gamma)^2, \frac{\delta}{SA}, \frac{\rho}{2S^2A^2})$ then $V_t(s) - V^{\pi_t}_{\theta_{K_t}}(s) \leq \epsilon$ for all time steps $t$ with probability at least $1 - \delta - \rho$.*

*Proof.* Let $\delta' = \frac{\delta}{SA}$. Because the sample complexity bound holds for each state–action pair with probability $\delta'$, the sample complexity bound holds overall by union bound with probability $1 - SA\delta' = 1 - \delta$.

For the remainder of the lemma, it suffices to show that the mean MDP has low error in known states. Let $A = \frac{1}{4}\epsilon(1-\gamma)^2$. In the known states, $\hat{R}_b(s,a) < A$; therefore $\eta_R(s,a,b) < A$ and $\eta_T(s,a,b) < \frac{(1-\gamma)A}{\gamma S}$. These imply $|R_\theta(s,a) - R_b(s,a)| < A$ and $\|P_\theta(\cdot|s,a) - P_b(\cdot|s,a)\|_\infty < \frac{(1-\gamma)A}{\gamma S}$. For the transition bound, we can convert the max norm bound into an $L^1$ norm by multiplying by $S$: $\|P_\theta(\cdot|s,a) - P_b(\cdot|s,a)\|_1 < \frac{(1-\gamma)A}{\gamma}$. It can be shown by Lemma 1 in Strehl and Littman (2008) that these bounds on the reward and transition errors lead to at most $|V_t(s) - V^{\pi_t}_{\theta_{K_t}}(s)| < \frac{2A}{(1-\gamma)^2} = \frac{\epsilon}{2}$ error, if using the mean reward function in known states. Because Algorithm 1 uses the internal reward function even in known states, there is an additional error less than $\frac{A}{1-\gamma}$. However, $\frac{A}{1-\gamma} + \frac{\epsilon}{2} < \epsilon$. □

*Proof of Theorem.* (Sketch) This theorem is a straightforward application of Proposition 1 in Strehl et al. (2006). The theorem requires three conditions: (1) Optimism: $V_t(s) \geq V^*(s) - \epsilon$. This was shown in Lemma 3 to hold with probability $1 - \delta$. (2) Accuracy: $V_t(s) - V^{\pi_t}_{M_{K_t}}(s) \leq \epsilon$. This was shown in Lemma 7 to hold with probability $1 - 2\delta$. (3) Sample complexity: The number of escape events is bounded by $\sum_{s,a} C(s,a)$. These three conditions are sufficient to prove the stated bound. □